%% file: main.tex
\documentclass[10pt]{article} 

\usepackage[accepted]{rlj} 

\usepackage{amssymb}            
\usepackage{mathtools}          
\usepackage{mathrsfs}           
\usepackage{graphicx}           
\usepackage{subcaption}         
\usepackage[space]{grffile}     
\usepackage{url}                
\usepackage{lipsum}             

\title{Collaboration Promotes Group Resilience in Multi-Agent RL}

\setrunningtitle{Collaboration for Resilience}

\author{
  Ilai Shraga\textsuperscript{1},
  Guy Azran\textsuperscript{1},
  Matthias Gerstgrasser\textsuperscript{2},
  Ofir Abu\textsuperscript{3},
  Jeffrey S. Rosenschein\textsuperscript{3},
  Sarah Keren\textsuperscript{1}
}

\emails{
  ilaishraga@campus.technion.ac.il,\,
  guy.azran@campus.technion.ac.il,\,
  matthias@seas.harvard.edu,\,
  ofirabu@cs.huji.ac.il,\,
  jeff@cs.huji.ac.il,\,
  sarahk@technion.ac.il
}

\affiliations{
$^{1}$\textbf{Taub Faculty of Computer Science, Technion - Israel Institute of Technology.}\\
$^{2}$\textbf{School of Engineering and Applied Sciences, Harvard University.}\\
$^{3}$\textbf{School of Computer Science and Engineering, The Hebrew University of Jerusalem.}
}

\input{content/contribution}

\keywords{Multi-Agent Reinforcement Learning, Group Resilience, Collaboration, Deep Reinforcement Learning.} 

\input{content/summary}
\input{comments}

\begin{document}

\makeCover  
\maketitle  

\input{content/abstract}

\newtheorem{example}{Example}

\input{macros}
\input{content/intro}
\input{content/background}
\input{content/model}
\input{content/method}

\input{content/empirical}

\input{content/conclusion}


\bibliography{main}
\bibliographystyle{rlj}

\end{document}

%% file: content/contribution.tex
\contribution{
    We introduce a novel definition of group resilience and formalize this notion, which corresponds to the group’s ability to adapt to unexpected changes.
    }
    {
    Prior work primarily focused on resilience in single-agent settings or adversarial MARL scenarios without a unified resilience measure.
    }

\contribution{
    We demonstrate that collaboration promotes group resilience, providing empirical evidence across multiple MARL benchmarks.
    }
    {
     Since prior work has mostly explored single-agent or adversarial settings, collaboration and its effects have not been considered.
    }

%% file: content/summary.tex
\summary{
To effectively operate in various dynamic scenarios, RL agents must be resilient to unexpected changes in their environment. Previous work on this form of resilience has focused on single-agent settings. In this work, we introduce and formalize a multi-agent variant of resilience, which we term {\em group resilience}. We further hypothesize that collaboration with other agents is key to achieving group resilience; collaborating agents adapt better to environmental perturbations in multi-agent reinforcement learning (MARL) settings. We test our hypothesis empirically by evaluating different collaboration protocols and examining their effect on group resilience. 
Our experiments show that all the examined collaborative approaches achieve higher group resilience than their non-collaborative counterparts. 
}

%% file: comments.tex
\newcount\Comments
\usepackage{xcolor}
\definecolor{darkgreen}{rgb}{0,0.7,0}

\newcommand{\kibitz}[2]{\ifnum\Comments=1{\color{#1}{#2}}\fi}

%% file: content/abstract.tex
\begin{abstract}

To effectively operate in various dynamic scenarios, RL agents must be resilient to unexpected changes in their environment. Previous work on this form of resilience has focused on single-agent settings. In this work, we introduce and formalize a multi-agent variant of resilience, which we term {\em group resilience}. We further hypothesize that collaboration with other agents is key to achieving group resilience; collaborating agents adapt better to environmental perturbations in multi-agent reinforcement learning (MARL) settings. We test our hypothesis empirically by evaluating different collaboration protocols and examining their effect on group resilience. 
Our experiments show that all the examined collaborative approaches achieve higher group resilience than their non-collaborative counterparts. 

\end{abstract}

%% file: macros.tex
\newtheorem{lemma}{Lemma}
\newtheorem{proposition}{Proposition}
\newtheorem{definition}{Definition}
\newtheorem{notation}{Notation}
\newtheorem{corollary}{Corollary}

\newcommand{\states}{S}
\newcommand{\initState}{s_{0}}
\newcommand{\actions}{A}
\newcommand{\rewardFunc}{R}
\newcommand{\transitionFunc}{P}
\newcommand{\discountFactor}{\gamma}
\newcommand{\utilityFunc}{U}
\newcommand{\utilityFuncGroup}{\mathcal{U}}

\newcommand{\terminalStates}{s_{end}}
\newcommand{\obsTokens}{\mathcal{O}}
\newcommand{\obsFunc}{O}
\newcommand{\jointActions}{\mathcal{A}}
\newcommand{\jointRewardFunc}{\mathcal{R}}
\newcommand{\jointTransitionFunc}{\mathcal{T}}

\newcommand{\commSym}{m}

\newcommand{\MDP}{M}
\newcommand{\MDPS}{\mathcal{M}}
\newcommand{\distMeasure}{\delta}
\newcommand{\Exp}{\mathbb{E}}
\newcommand{\MDPdist}{\Psi}

\newcommand{\agent}{N} 
\newcommand{\numAgents}{n}
\newcommand{\policy}{\pi}
\newcommand{\policies}{\Pi}

\newcommand{\RL}{RL}
\newcommand{\allMDPs}{\mathcal{M}}

\newcommand{\atomicPerturbation}{\phi}
\newcommand{\perturbationFunc}{\Phi}

\newcommand{\ResRL}{Resilience RL} 

\newcommand{\pertVol}{k}
\newcommand{\pertBound}{K}
\newcommand{\confLevel}{J}
\newcommand{\msgLen}{m_{l}}
\newcommand{\msgFreq}{m_{f}}
\newcommand{\pertFreq}{t_{pert}}
\newcommand{\obsElim}{\omega}
\newcommand{\resLevel}{C}
\newcommand{\colabProt}{\theta}
\newcommand{\colabProts}{\Theta}
\newcommand{\mapSize}{6}
\newcommand{\numTaxis}{2}
\newcommand{\numPassengers}{2}
\newcommand{\numPassLocs}{4}
\newcommand{\totalTSteps}{T}
\newcommand{\statesDist}{d}
\newcommand{\immediateR}{r}
\newcommand{\numExperiments}{10}

\newcommand{\RelativeToOpt
}{Relative to Optimum}

\newcommand{\RelativeToOrigin
}{Relative to Origin}

\newcommand{\transition}{\tau}

\newcommand{\qvalfunc}{Q}

%% file: content/intro.tex
\section{Introduction}

Reinforcement Learning (RL) agents are typically required to operate in dynamic environments and must develop an ability to quickly adapt to change.
Promoting this ability is hard, even in single-agent settings
\citep{padakandla_survey_2022}.
When the RL agent operates alone, it needs to adapt its behavior to the changing and possibly partially observable environment.
For a group, this is even more challenging. In addition to the dynamic nature of the environment, agents need to deal with high variance caused by the changing behavior of other agents.
Nevertheless, the ability of autonomous agents, individually or as a group, to adapt to environmental changes is highly desirable in real-world settings where dynamic environments are the norm. Therefore, if a group is to reliably pursue its objective, it should be able to handle unexpected environmental perturbations.



Many recent Multi-Agent RL (MARL) studies show that collaboration boosts agents’ performance \citep{christianosSharedExperienceActorCritic2020,foerster_learning_2016,honhaga2024simulation,jaques_social_2019,lowe_multi-agent_2020,qianSurveyReinforcementLearning2019,xu_url_2012}. We aim to understand how collaboration promotes the {\em group resilience}, i.e., how collaboration affects the ability of a group to rapidly regain a significant portion of their prior performance after unexpected perturbations occur in their environment.  
We therefore investigate whether agents that collaborate in different ways indeed reclaim a larger share of their former performance after a disturbance occurs.

Contrary to investigations of \textit{transfer learning}~\citep{liang_parallel_2020, zhu_transfer_2023} or \textit{curriculum learning}~\citep{portelas_automatic_2020}, we do not assume a stationary target domain in which agents will be deployed, nor do we have a dedicated training phase to prepare agents for various environments. Moreover, unlike typical generalisation scenarios, which evaluate final performance on previously unseen tasks, our resilience metric explicitly captures how group performance recovers following random but bounded perturbations. We focus on multi-agent reinforcement learning (MARL) settings, for which previous work demonstrates how collaboration enables groups to operate efficiently in complex but stationary environments~\citep{christianosSharedExperienceActorCritic2020,foerster_learning_2016,jaques_social_2019}. We offer empirical evidence that collaboration promotes resilience in non-stationary environments, facilitating it via existing~\citep{jaques_social_2019} and custom communication protocols.

The literature offers a range of definitions for resilience in both single and multi-agent settings~\citep{pattanaik_robust_2017,phan_learning_2020,vinitsky_robust_2020,zhangResilientRobotsConcept2017}, primarily focusing on resilience in the face of adversarial behavior \citep{saulnier_resilient_2017,phan_learning_2020}, and on algorithms for training resilient agents that define resilience in a model-specific way. However, these frameworks do not define resilience in a unified, measurable way. 
We concentrate on non-adversarial settings, emphasizing the agents' ability to enhance resilience through collaboration. We define and measure group resilience based on agents' performance under unexpected and random environmental perturbations of bounded magnitude. 
Our goal is to provide a unified, general measure of resilience that is based on a user-specified distance metric between environments and that is adaptable to various settings. The perturbations we model represent unexpected changes in the real world, reflecting practical scenarios that agents might encounter.

\begin{figure}[ht]
     \centering
     \begin{subfigure}[b]{0.4\textwidth}
         \centering
       \includegraphics[width=\textwidth]{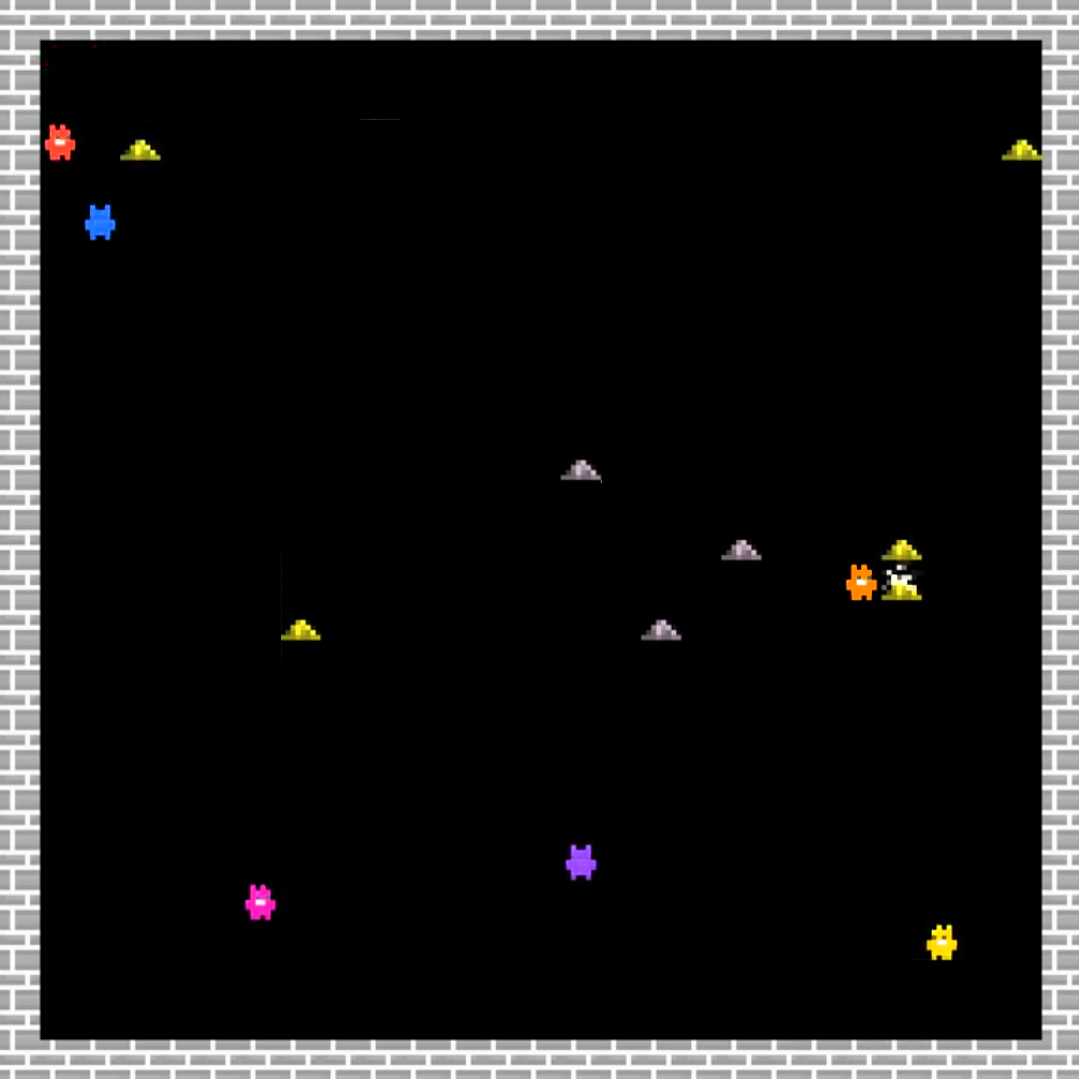}
         \caption{}
         \label{fig:coop-mining-example:clean}
     \end{subfigure}
     \hspace{1cm}
     \begin{subfigure}[b]{0.4\textwidth}
         \centering
         \includegraphics[width=\textwidth]{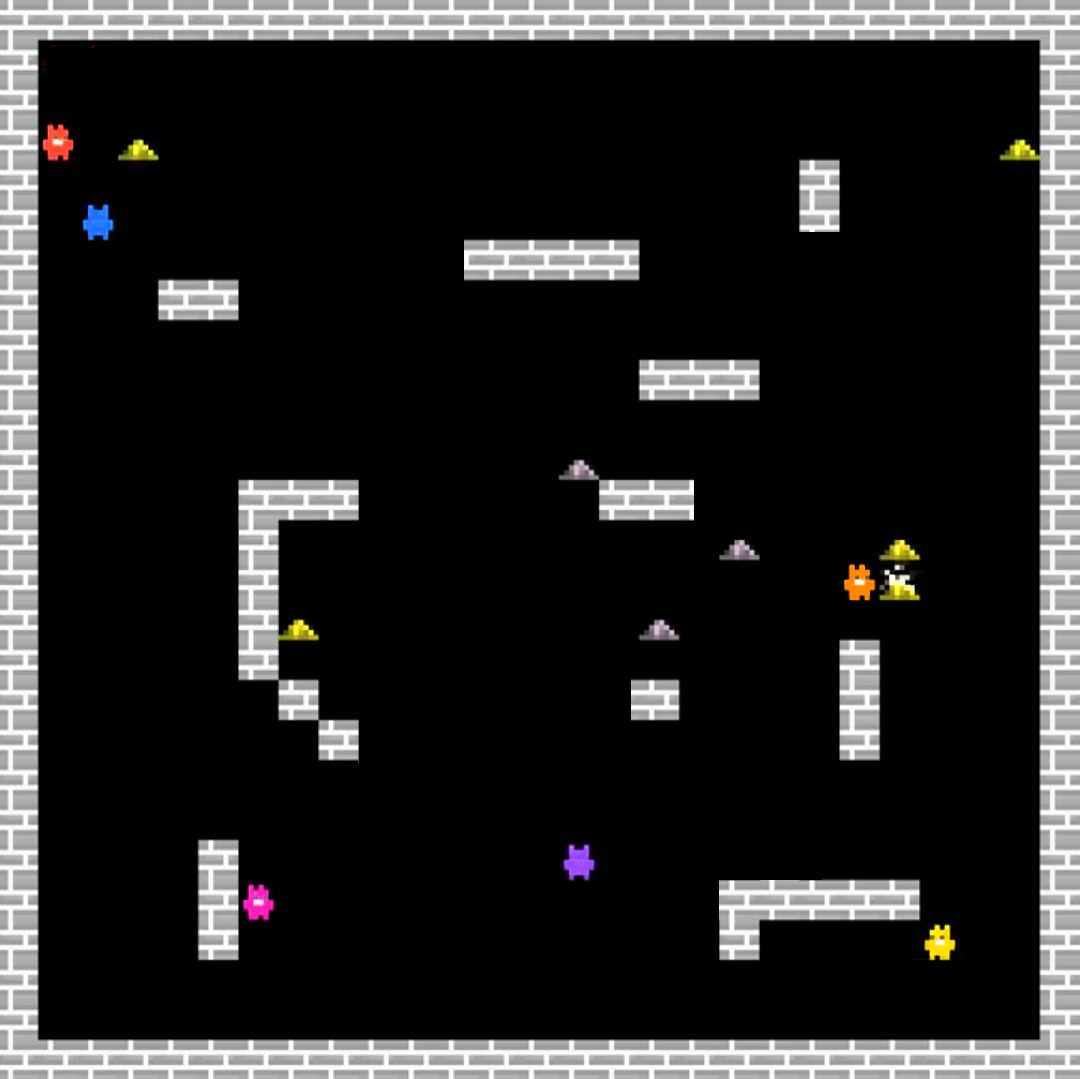}
         \caption{}
         \label{fig:coop-mining-example:pert}
     \end{subfigure}
     \hfill
        \caption{An illustration of the coop-mining domain. There are 5 miner agents (colorful creatures). Two types of ores (resources) appear randomly on the map: iron (grey mounds) and gold (yellow mounds). \ref{fig:coop-mining-example:clean} shows a clean mine with walls only on the boundaries. \ref{fig:coop-mining-example:pert} shows a perturbation of the environment; perturbations are newly introduced non-traversable walls (solid lines).}
         \label{fig:coop-mining-example}
\end{figure}

\begin{example}\label{example}
Figure~\ref{fig:coop-mining-example:clean} depicts a multi-agent variation of the coop-mining domain \citep{leibo2021meltingpot}.
All miners are employed by the same mining company, which aims to maximize the group’s total revenue.
Miners may find two types of ores, iron and gold, anywhere in the mine. A single miner can extract one unit of iron alone, which sells for \$100. Mining gold requires two miners to strike the ore multiple times in unison, but it sells for \$800.
As the miners excavate, unstable terrain or mild earthquakes may cause cave-ins that create unexpected blockages in random locations within the mine (see Figure~\ref{fig:coop-mining-example:pert}). At this point, the miners may benefit from sharing information about these blockages to enhance the group's resilience, i.e., their ability to adapt to the perturbations.
We aim to show that agents trained to communicate prior to perturbations are more resilient to unexpected disruptions and can more quickly adapt to the changing mine layout.
\end{example}

Our contributions are threefold.
First, we suggest a \textbf{new unified measure} called \emph{group resilience}, which corresponds to the group's ability to adapt to unexpected changes. This formulation covers a wide range of real-world multi-agent problems. Second, we design and implement a MARL framework for testing multi-agent resilience in the face of environment perturbations. Finally, we provide empirical evidence supporting our hypothesis that collaboration promotes resilience in MARL.

%% file: content/background.tex
\section{Background}

\textbf{\textit{Reinforcement learning} (RL)} is a learning paradigm in which agents learn by observing the world, acting within it, and receiving rewards (positive or negative) for achieving certain states or state transitions. RL problems commonly model the world as a \emph{Markov decision process} (MDP) \citep{bellmanMarkovianDecisionProcess1957} $\MDP = \langle\states, \actions, \rewardFunc, \transitionFunc, \discountFactor\rangle$ where \(\states\) is a set of possible states, \(\actions\) is a set of agent actions,
\(\transitionFunc:\states\times\actions\times\states \rightarrow [0, 1]\)
is the state transition function, \(\rewardFunc:\states\times\actions\times\states \rightarrow \mathbb{R}\) is the reward function, and \(\discountFactor\) is the reward discount factor.
The objective is to find a policy \(\pi^*\) such that \(\pi^* \in \arg\max_{\pi}\mathbb{E}[J(\pi)]\), where:
$J(\policy) 
  = \mathbb{E}_{s_t,\,s_{t+1}\sim\transitionFunc;\;a_{t}\sim\policy}
    \!\Biggl[\sum_{t=0}^\infty \discountFactor^t \,\rewardFunc\bigl(s_t, a_t, s_{t+1}\bigr)\Biggr]
$
is the expected return of policy \(\policy\). 




\textbf{\textit{Multi-Agent Reinforcement Learning} (MARL)} extends RL to settings with multiple agents.
We model the environment as a \emph{Markov Game} (MG)~\citep{littmanMarkovGamesFramework1994a}, in which each agent selects an individual action, receives a reward, and the transition to the next state depends on the joint action of all agents.
We denote by $\utilityFuncGroup_\pi(M)$ the cumulative discounted return obtained by a joint policy~$\pi$ in environment~$M$. Since we focus on settings in which the environment might change, we define the environment-specific optimum as
\[
  \utilityFuncGroup^{\max}(M)=\sup_{\pi} \utilityFuncGroup_\pi(M)\;
\]
We follow 
\cite{jordanScoreNormalizationMultiEnv2020}
in using the \emph{normalised utility} $\hat{\utilityFuncGroup}(M)=\utilityFuncGroup(M)/\utilityFuncGroup^{\max}(M)\in[0,1]$, so scores remain comparable even when perturbations affect rewards or dynamics.


%% file: content/model.tex
\section{Measuring Group Resilience}
\label{sec:measuring}

We aim to promote the ability of a group to adapt to random perturbations in their environment. We refer to this ability as {\em group resilience} and formally define it below. Our definition explicitly quantifies the group's adaptive capacity, i.e.,  the extent of performance recovery following unexpected perturbations. We will then explore our hypothesis on promoting group resilience by facilitating collaboration.

To measure group resilience, we take inspiration from the field of multi-agent robotics. Specifically, the work of~\cite{saulnier_resilient_2017} aims to produce a control policy that allows a team of mobile robots to achieve desired performance in the presence of faults and attacks on individual group members.
Accordingly, a group of robots achieves \emph{resilient consensus} if the cooperative robots' performance remains in some desired range, even in the presence of a bounded number of non-cooperative robots.

Similarly, we define \emph{group resilience} as the ability of agents to maintain at least a fixed fraction of their original performance when the environment undergoes an unexpected perturbation bounded in magnitude. Unlike resilience to adversarial agents, our definition explicitly addresses changes in the environment itself, implicitly accounting for the corresponding changes in agents' observations and experiences. Formally, our definition relies on (i) a distance measure $\distMeasure(\MDP, \MDP')$ quantifying the magnitude of change between an original MDP $\MDP$ and a perturbed MDP $\MDP'$, and (ii) the utility measure that quantifies group performance. 
Given these measures, we require that any perturbed environment within a bounded distance $\pertBound$ from the original environment results in decreased performance by at most a constant factor $C_\pertBound$\footnote{Note the similarity to the classical $\epsilon$-$\delta$ continuity definition.}.

\begin{figure}[ht]
\centering
\begin{minipage}{0.18\textwidth}
  \centering
  \includegraphics[width=1\textwidth]{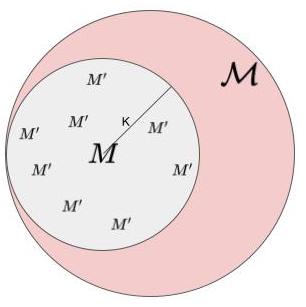}
\end{minipage}%
\hspace{0.7cm}
\begin{minipage}{0.65\textwidth}
  \raggedright
  \caption{Resilience considers performance differences for perturbed environments $\MDP'$ within some distance from a reference environment $\MDP$.}
  \label{fig:res-def}
\end{minipage}
\end{figure}

We note that a variety of subtly different formal definitions can satisfy this intuitive requirement. We provide here only those definitions directly relevant to our experiments. Specifically, the following definitions assume that a designer aims to guarantee resilience over a subset of perturbed environments within a specified distance from a reference MDP $\MDP$ (see Figure~\ref{fig:res-def}). 

\vspace{-13mm}

An important issue to consider is that once an environment is perturbed, the optimal utility that can be achieved might change. Thus, in principle, it might be beneficial to consider the performance level of a team with regard to the normalized utility $\hat{\utilityFuncGroup}$ in each environment as follows.  

\begin{definition}[Relative-to-optimum $\resLevel_\pertBound$-resilience]
\label{def:pert_group_opt}
Given a class of MDPs $\MDPS$, a reference MDP $\MDP \in \MDPS$, and a bound $\pertBound \in \mathbb{R}$, we say a group of agents is {\em $\resLevel_\pertBound$-resilient} over $\MDPS$ relative to optimum if
\[
\forall \MDP' \in \MDPS : \distMeasure(\MDP, \MDP') \leq \pertBound \Longrightarrow
\hat{\utilityFuncGroup}(\MDP') \geq \resLevel_\pertBound \cdot \hat{\utilityFuncGroup}(\MDP)\;
\]
\end{definition}
Setting $\resLevel_\pertBound \in [0, 1]$ ensures performance degradation remains bounded. 

    In some settings, it may be impossible to know or approximate the optimal utility. Nevertheless, if the goal is to compare the resilience of multiple groups, it is sufficient to compare their utilities in the original and perturbed environments without normalization.

\begin{definition}[Relative-to-origin $\resLevel_\pertBound$-resilience]
\label{def:pert_group}
Given a class of MDPs $\MDPS$, a reference MDP $\MDP \in \MDPS$, and a bound $\pertBound \in \mathbb{R}$, we say a group of agents is {\em $\resLevel_\pertBound$-resilient} over $\MDPS$ relative to origin $\MDP$ if
\[
\forall \MDP' \in \MDPS : \distMeasure(\MDP, \MDP') \leq \pertBound \Longrightarrow
\utilityFuncGroup(\MDP') \geq \resLevel_\pertBound \cdot \utilityFuncGroup(\MDP)\;
\]
\end{definition}
The above definition is especially relevant to our aim of comparing the performance of groups that undergo the same perturbation but adopt different collaboration approaches. 


While resilience over all $\MDP'$ within distance $\pertBound$ provides a strong guarantee, it can be impractical. Thus, given a probability distribution $\MDPdist$ over $\MDPS$, we define resilience-in-expectation as follows:

\begin{definition}[Relative-to-origin $\resLevel_\pertBound$-resilience in expectation]
\label{def:expres}
Given an MDP $\MDP$, a distribution $\MDPdist$ over a class of MDPs, and a bound $\pertBound \in \mathbb{R}$, a group of agents is {\em $\resLevel_\pertBound$-resilient in expectation} over $\MDPdist$ relative to origin $\MDP$ if
\[
\mathbb{E}_{\MDP'\sim \MDPdist\mid\distMeasure(\MDP, \MDP')\leq \pertBound}\left[\utilityFuncGroup(\MDP')\right] \geq \resLevel_\pertBound \cdot \utilityFuncGroup(\MDP)\;
\]
\end{definition}

This definition guarantees expected group performance remains above a threshold when averaging over perturbed environments within distance $\pertBound$ from the reference environment. 

Polynomially many samples from $\MDPdist$ suffice to approximate this expectation with high probability. Formally, if we assume that $\utilityFuncGroup(\MDP')$ is a random variable with finite variance $\sigma^2$ drawn i.i.d. from $\MDPdist$ under the condition $\distMeasure(\MDP,\MDP')\leq \pertBound$, then by Chebyshev's inequality, the expected normalized utility can be approximated within distance $\epsilon$ of the true expectation with probability at least $\delta$ using $\frac{\sigma^2}{\epsilon^2(1-\delta)}$ samples.

Note that Definitions~\ref{def:pert_group} and~\ref{def:expres} compare agents' performance under perturbation directly against their performance in the unperturbed environment, without explicitly considering baseline performance. Hence, a non-optimal baseline policy (e.g., no-op actions) could misleadingly indicate high resilience. Depending on analysis objectives, our measures may thus require complementary baseline normalization.

\subsection{Perturbations}
In this work, we are interested in settings in which we have an initial environment and a set of \emph{perturbations} that can occur. A perturbation \(\atomicPerturbation: \MDPS \mapsto \MDPS\) is a function transforming a source MDP into a modified MDP.
An \emph{atomic perturbation} is a perturbation that changes only one of the basic elements of the original MDP. Given an MDP
\(\MDP = \langle\states, \actions, \rewardFunc, \transitionFunc, \discountFactor\rangle\)
and perturbation \(\atomicPerturbation\), we denote the resulting MDP after applying \(\atomicPerturbation\) by
\(\MDP^{\atomicPerturbation} = \langle\states^{\atomicPerturbation}, \actions^{\atomicPerturbation}, \rewardFunc^{\atomicPerturbation}, \transitionFunc^{\atomicPerturbation}, \discountFactor^{\atomicPerturbation}\rangle\).

Among the variety of perturbations that may occur, we focus here on three types of atomic perturbations. \emph{Transition function perturbations} modify the distribution over the next states for a single state-action pair. \emph{Reward function perturbations} modify the reward of a single state-action pair. \emph{Initial state perturbations} change the initial state of the MDP.

\begin{definition}[Transition Function Perturbation]
A perturbation \(\atomicPerturbation\) is
a \emph{transition function perturbation} if for every 
MDP \(\MDP = \langle\states, \actions, \rewardFunc, \transitionFunc, \discountFactor\rangle\), 
\(\MDP^{\atomicPerturbation}\) is identical to \(\MDP\) except that for a single action-state pair 
\(s\in \states\) and \(a\in \actions\), 
\({\mathbb{P}}_s^a[\states]\neq {\mathbb{P}}_s^{a,\atomicPerturbation}[\states]\).
\end{definition}

\begin{definition}[Reward Function Perturbation]
A perturbation \(\atomicPerturbation\) is
a {\em reward function perturbation} if for every 
MDP \(\MDP = \langle\states, \actions, \rewardFunc, \transitionFunc, \discountFactor\rangle\), 
\(\MDP^{\atomicPerturbation}\) is identical to \(\MDP\) except that for a single action-state pair 
\(s\in \states\) and \(a\in \actions\), 
\(\immediateR_s^a \neq {\immediateR}_s^{a,\atomicPerturbation}\).
\end{definition}

\begin{definition}[Initial State Perturbation]
A perturbation \(\atomicPerturbation\) is
an {\em initial state perturbation} if for every 
MDP \(\MDP = \langle\states, \initState, \actions, \rewardFunc, \transitionFunc, \discountFactor\rangle\), 
\(\MDP^{\atomicPerturbation}\) is identical to \(\MDP\) except that \(\initState \neq \initState^{\atomicPerturbation}\).
\end{definition}

\addtocounter{example}{-1}
\begin{example}[continued]
In our Coop-Mining domain, a path blockage can be modeled as an initial state perturbation or as a transition function perturbation that stops the agent from transitioning to the adjacent cell.
This is represented by changing \(\transitionFunc\) to express a probability distribution \({\mathbb{P}}^a_s[\states]\) that is set to be \({\mathbb{P}}^a_s(s') = 0\) when \((s,a,s')\) represents crossing the blocked area, and \({\mathbb{P}}^a_s(s) = 1\), which represents staying in the same place.
A change in an ore's (gold) location can be represented by two atomic perturbations: one that replaces the reward for mining at the original location with a negative reward, and one that adds a positive reward for mining at the new location.
\end{example}

A straightforward way to measure the distance $\distMeasure$ between two MDPs is to count the minimal number of atomic perturbations that transition the original MDP into the transformed one. While this metric has some limitations, it is sufficient for our experimental settings. Future work should consider more sophisticated distance measures between MDPs. For instance, one can define the overall distance based on a state-level distance $d(s,s')$, instantiated using established measures from the literature, such as Kantorovich (Wasserstein) distances~\cite{dobrushinPrescribingSystemRandom1970,songMeasuringDistanceFinite2016a}, bisimulation-based distances~\cite{fernsMetricsFiniteMarkovDecision2004}, or transfer-oriented distance metrics~\cite{ammarAutomatedMeasureMDPSimilarity2014}.

%% file: content/method.tex
\section{Facilitating Collaboration via Communication}
\label{sec:method}

Equipped with a measure for group resilience, we now focus on maximizing the resilience of a group of RL agents by facilitating collaboration. Recent MARL work suggests various approaches for promoting collaboration \citep{jaques_social_2019,mahajanMAVENMultiAgentVariational2019,rashid_qmix_2018}. In this work, we focus on communication \citep{christianosSharedExperienceActorCritic2020,foerster_learning_2016}.

To support collaboration, communication protocols produce messages that encode information valuable to other agents' learning. Prior literature in MARL explicitly defines effective collaboration as relying upon the exchange of meaningful information, allowing agents to coordinate their actions and adapt jointly to dynamic environmental changes~\citep{Foerster2016LearningTC,Jaques2019SocialIA}.
We examine different communication protocols based on broadcasting observations that are least aligned with their previous experiences.
Misalignment corresponds to the agents' familiarity with the environment, which may decrease due to perturbations. By communicating misaligned transitions, agents increase the other agents’ familiarity with the environment.

We present two definitions of misalignment. The first is taken from \citep{gerstgrasserSelectivelySharingExperiences2023}, and measures misalignment using the Temporal Difference (TD) error of a given observation. Formally, let \(e_t = \langle s_t,a_t,s_{t+1},r_t\rangle\) be the experience at time \(t\), that is, transitioning to state \(s_{t+1}\) after taking action \(a_t\) in \(s_t\) and receiving reward \(r_t\). Let \(\policy_p\) represent the policy of agent \(p\), and let \(Q^{\policy_p}\) represent the \(Q\)-function of policy \(\policy_p\), i.e., the expected value of taking action \(a\) in state \(s\) and following \(\pi\) thereafter. Given an experience, the {\em TD-error} is:

\begin{equation}
\mathrm{TD}(e_t) = \left|r_t + \gamma \,\max_{a}\,Q^{\pi_p}(s_{t+1}, a) - Q^{\pi_p}(s_t,\,a_t) \right|
\label{eq:td-error}
\end{equation}

This definition is inspired by Prioritized Experience Replay \citep[PER]{schaul_prioritized_2016}, according to which a deep Q-Network (DQN) agent \citep{mnihHumanlevelControlDeep2015} maintains a buffer of past transitions and prioritizes them in a way that expedites training.

A second measure of {\em misalignment} of a transition considers the difference between the observed reward \(r_t\) and the expected reward \(\hat{r_t}\). 
The misalignment for agent \(p\) at \(s_t\) after taking action \(a_t\), denoted by \(\confLevel_{s_t, a_t}^p\), is defined as
\begin{equation}
\confLevel_{s_t, a_t}^p 
\;=\; 
\frac{\bigl|r_t \;-\; \hat{r_t}\bigr|}{r_t}
\label{eq:misalignment-def}
\end{equation}
where \(\hat{r_t} \approx Q^{\pi_p}(s_t, a_t) - Q^{\pi_p}(s_{t+1}, \pi_p(s_{t+1}))\).


Using these two measures, we created different communication protocols in which misalignment is used to decide which messages to broadcast to other agents, described below:

\begin{enumerate}
    
    \item \textbf{No Communication} – Agents do not share information (used as a baseline). 

    \item \textbf{Mandatory Broadcast} – Each agent \(p\) broadcasts at state \(s_t\) its most misaligned experiences, i.e., transitions with the highest \(\confLevel_{s_t,a_t}^p\). A message consists of the misaligned transitions \(\transition\). Messages are received by all other agents and inserted into their replay buffers.
    The number of transitions broadcast at each time step is bounded by the channel bandwidth parameter \(\msgLen\).

    \item \textbf{Emergent Communication} – Each agent \(p\) broadcasts a discrete communication symbol \(\commSym^p_t\) among a given set of symbols, at each state \(s_t\). Individual messages of all agents are concatenated into a single vector \(\commSym_t = [\commSym^1_t \dots \commSym^N_t]\), which is included as an additional observation signal that all agents receive at the next time step \((t+1)\).  We distinguish between two sub-cases: Self-Centric and Global-Centric. In  \textbf{Emergent Self-Centric Communication}, each agent \(p\) uses counterfactual reasoning and chooses a symbol \(\commSym_t^p\) that would have minimized \emph{its own} misalignment at time step \(t-1\). Formally, the loss function for \(\policy_\commSym\) is:
    \[
    L_t 
    \;=\; 
    \Bigl|\arg \min_{\commSym}\,\confLevel_{s_{t-1}, a_{t-1}}^{p_m} 
           \;-\; \confLevel_{s_{t-1}, a_{t-1}}^p
    \Bigr|
    \]
    where \(\confLevel_{s_{t-1}, a_{t-1}}^{p_m}\) is the misalignment at \((t-1)\) had it received message \(\commSym\).
    In \textbf{Emergent Global-Centric Communication}, agents observe the misaligned observations of all other agents at each time step.
    Each agent \(p\) is rewarded for choosing a symbol \(\commSym_t^{p}\) that would have minimized the \emph{total} misalignment of the group at \((t-1)\).
    Agents maintain a model that predicts the global (average) misalignment of the other agents given an observation and messages.
    Formally:
    \[
    L_t 
    \;=\; 
    \Bigl|\arg \min_{\commSym}\,\hat{\confLevel}_{s_{t-1}, a_{t-1}}^{P_m} 
           \;-\; \confLevel_{s_{t-1}, a_{t-1}}^P
    \Bigr|
    \]
    where \(\hat{\confLevel}_{s_{t-1}, a_{t-1}}^{P_m}\) is the counterfactual predicted average misalignment of the other agents, having received \(\commSym\), and \(\confLevel_{s_{t-1}, a_{t-1}}^{P}\) is the actual average misalignment of the other agents.
    \item \textbf{suPER -- Selectively Sharing Experiences} \citep{gerstgrasserSelectivelySharingExperiences2023}:  
    Each agent \(p\) broadcasts at state \(s_t\) its experiences with the highest TD-errors (Equation~\ref{eq:td-error}). These transitions are inserted into the replay buffers of receiving agents. The number of broadcast transitions per time step is bounded by \(\msgLen\). suPER leverages PER insights so that not all experiences are equally relevant for learning, and thus it supports decentralized training with minimal communication overhead, compatible with standard DQN \citep{mnihHumanlevelControlDeep2015} variants.

\end{enumerate}

%% file: content/empirical.tex
\section{Empirical Evaluation}

Our empirical evaluation aims to assess the effect collaboration has on group resilience. Specifically, we measure and compare the utility of groups of agents in randomly perturbed environments (various types of atomic perturbations), where each group implements a different communication protocol and learning approach\footnote{Our code base and full results are available at \url{https://github.com/CLAIR-LAB-TECHNION/CollaborationForResilience}}. We conducted two sets of experiments. The first uses our custom communication protocols in environments requiring relatively simple forms of collaboration.
The second uses SOTA communication in an abstraction of Experiment \ref{example}. Experiments ran on a \texttt{x86\_64} CPU running Ubuntu 20.04.6.

\paragraph{Experiment 1: Simple Collaboration} \label{sec:experiment1}

We experiment with three multi-agent RL environments. \textbf{Cleanup} \citep{SSDOpenSource}: Seven agents must balance harvesting apples (individual rewards) and cleaning a river (enables regrowth but prevents harvesting). Agents can fine each other, and each observes a raw image of its surroundings. \textbf{Harvest} \citep{SSDOpenSource}: Similar to Cleanup, but apple regrowth depends on proximity rather than a river, requiring coordinated harvesting to avoid depletion. \textbf{Multi-Taxi} \citep{azranContextualPreplanningReward2024}: Taxis transport passengers in a configurable grid world with perturbations. Observations are symbolic state vectors. Rewards include high positive for drop-offs, small negative for time steps, and large negative for collisions. Grid sizes range from $5\times5$ to $8\times8$ with 2–3 taxis and passengers.

We trained individual neural networks for every RL agent using the distributed Asynchronous Advantage Actor-Critic (A3C) algorithm \citep{mnih_asynchronous_2016}. For the Multi-Taxi domain, agents were implemented using a Deep Q-Network \citep{mnihHumanlevelControlDeep2015}.
We evaluate a spectrum of communication protocols within our framework. The \textbf{No Communication} protocol involves agents operating independently without exchanging information. In the \textbf{Social Influence} protocol, as suggested by \citep{jaques_social_2019}, agents broadcast messages aimed at maximizing their impact on the immediate behaviors of other agents. The \textbf{Mandatory Communication} protocol requires agents to share their top \(\msgLen\) most misaligned transitions, as detailed in Section~\ref{sec:method}. In the \textbf{Emergent Self-Centric Communication} protocol, agents broadcast a discrete symbol at each step that would have minimized their previous step's misalignment. In the \textbf{Emergent Global-Centric Communication} protocol, agents monitor the current misalignment levels of their peers and broadcast symbols that aim to minimize the group's overall misalignment.

In the Cleanup and Multi-taxi domains, we used two types of perturbations. The first is a transition function perturbation, randomly adding non-traversable obstacles (e.g., walls) to the map. The second is an initial state perturbation, randomly changing the initial configuration (e.g., changing the river location in Cleanup, or initial taxi/passenger locations in Multi-taxi). In Harvest and Multi-taxi, we used a reward function perturbation, randomly reallocating rewards/resources (e.g., eliminating passengers or apples). We measure the perturbation's magnitude using the state-distance approach of \citep{songMeasuringDistanceFinite2016a} described in Section~\ref{sec:measuring}. We experiment with bounds \(\pertBound\in\{50,150,200\}\). For each initial environment \(\MDP\) and bound \(\pertBound\), we uniformly sample from possible perturbed \(\MDP'\) such that \(\distMeasure(\MDP,\MDP') \le \pertBound\), applying random atomic perturbations until the desired magnitude is reached.

To measure the effect of perturbations on group performance, we calculate the average utility throughout training before and after perturbation. We repeat each experiment 8 times with different random seeds. The process described above is used to generate perturbations for each seed. To measure resilience, we use Definition~\ref{def:expres} with a uniform distribution \(\Psi\). We let
\(
C_K 
= \tfrac{\mathrm{avg}\bigl(\utilityFuncGroup(\MDP')\bigr)}{\utilityFuncGroup(\MDP)},
\)
where \(\MDP'\) is drawn from \(\Psi\) within distance \(\pertBound\).

\paragraph{Experiment 2: Cooperative Mining} \label{sec:experiment2}
Here, we examine the \textbf{Coop-Mining} domain, designed to test various social interactions. There are five agents, and, as described in Example~\ref{example}, the domain incentivizes coordinating to gather resources, balancing reliable low-reward iron versus high-yield gold that requires cooperation.

We trained agents similarly to Experiment~1, but using Deep Q-Networks (DQNs) \citep{mnihHumanlevelControlDeep2015}.
We employ the \textbf{suPER} advanced communication protocol with the hyperparameters in the original work, whereby agents transmit their highest TD-error experiences into others' replay buffers, leveraging PER ideas for decentralized training with minimal overhead.

We apply transition function perturbations of varying magnitudes, measuring distance similarly to Experiment~1. Resilience and utility are calculated likewise.

%% file: content/conclusion.tex
\subsection{Results}

Table~\ref{table:c_levels} shows results for Experiment 1, reporting mean \(\resLevel_\pertBound\)-resilience with perturbations of varying magnitudes, alongside \(\utilityFuncGroup\) in the non-perturbed environment (std.\ dev.\ in parentheses). Figure~\ref{fig:cleanup-compare} compares group resilience in the Cleanup domain across different communication protocols, grouped by perturbation intensity, while Figure~\ref{fig:cleanup-training} shows the utility throughout training for \(\pertBound=200\). Figures~\ref{fig:coop-compare} and \ref{fig:coop-training} similarly present results for Experiment 2 (Coop-Mining).

\input{content/results-table}

\input{content/results-graphs4}

\input{content/results-graphs5}
In both experiments, we observe that all collaborative approaches achieve higher resilience than the no-communication approach, supporting our main hypothesis. The effect is more pronounced for larger-magnitude perturbations (e.g., a small 3\% increase in resilience with \(K=50\) in Cleanup versus a 180\% increase with \(K=200\)). We also observe that the global-centric approach generally outperforms the self-centric approach (higher or similar resilience, and higher initial performance). This further reinforces that collaboration promotes resilience in that agents can recover after a perturbation a larger fraction of their previous performance, even if they are self-interested.

\vspace{-1.5mm}
\section{Conclusion}

We suggest that collaboration promotes resilience: we hypothesized that agents who learn to collaborate will adapt more quickly to changes in their environment.    
In support of this agenda, we introduced a novel formulation for group resilience. To the best of our knowledge, this is the first measurement of group resilience that is relevant to MARL settings.
In addition, we presented an empirical evaluation of various MARL settings and communication protocols that show that collaboration via communication can significantly increase resilience to changing environments.

While we examined our approach in MARL settings with homogeneous agents that collaborate via communication, we intend to examine additional methods for collaboration in settings with heterogeneous groups of agents as a next step. Additionally, we intend to explore resilience in real-world domains, including multi-robot settings.

It is noteworthy that the recent global pandemic perturbed many aspects of the environments in which we operate. In such cases, people used to certain kinds of collaboration before the pandemic may have found it easier to adjust to the unfamiliar constraints that were imposed. We believe our results reflect a quite specific benefit that automated agents can derive from collaborating with one another. We do note that many usual caveats on AI research apply, especially concerning tasks that might not be of societal benefit. We leave this for future work, noting potential solutions in existing research on differential privacy and federated learning.
\clearpage

%% file: content/results-table.tex
\begin{table*}[htbp]
    \caption{Average (and standard deviation) $\resLevel_{\pertBound}$-resilience for Cleanup, Harvest, and Multi-Taxi.}
    \vspace{-1.1mm}
    \begin{center}
        \scalebox{0.65}{
                    \begin{tabular}{||l||c|c|c|c||c|c|c|c||c|c|c|c||}
            \hline
            &\multicolumn{4}{c||}{\bf Cleanup}
            &\multicolumn{4}{c||}{\bf Harvest}
            &\multicolumn{4}{c||}{\bf Multi-Taxi} \\
            \hline
            &\multicolumn{1}{c|}{$\resLevel_{\pertBound = 50}$}
            &\multicolumn{1}{c|}{$\resLevel_{\pertBound = 150}$}
            &\multicolumn{1}{c|}{$\resLevel_{\pertBound = 200}$}      
            &\multicolumn{1}{c||}{$\utilityFuncGroup$}
            &\multicolumn{1}{c|}{$\resLevel_{\pertBound = 50}$}
            &\multicolumn{1}{c|}{$\resLevel_{\pertBound = 150}$}
            &\multicolumn{1}{c|}{$\resLevel_{\pertBound = 200}$}      
            &\multicolumn{1}{c||}{$\utilityFuncGroup$}
            &\multicolumn{1}{c|}{$\resLevel_{\pertBound = 50}$}
            &\multicolumn{1}{c|}{$\resLevel_{\pertBound = 150}$}
            &\multicolumn{1}{c|}{$\resLevel_{\pertBound = 200}$}      
            &\multicolumn{1}{c||}{$\utilityFuncGroup$}\\
            
            \hline      
            \hline
            
            \bf No com- & 0.62 & 0.21 & 0.06 & 3.56 & 0.64 & 0.38 & 0.25 & 128.18 
                          & 0.67 & 0.35 & 0.27 & 141.35 \\
            \bf munication & (0.25) & (0.14) & (0.05) & (1.32)
                          & (0.15) & (0.11) & (0.12) & (94.84) 
                          & (0.14) & (0.17) & (0.17) & (40.59)\\                  
            
            \hline
            \bf Social & {\bf 0.78} & 0.40 & 0.25 & 7.49 & 0.77 & 0.50 & 0.36 & 132.68 
                         & 0.69 & {\bf 0.51} & {\bf 0.38} & 149.45 \\ 
            \bf Influence & (0.17) & (0.17) & (0.13) & (2.59) 
                         & (0.13) & (0.13) & (0.11) & (100.51) 
                         & (0.10) & (0.17) & (0.13) & (45.18)\\     
            \hline
            \bf Mandatory & 0.69 & 0.40 & 0.21 & 4.47 
                          & 0.72 & 0.48 & {\bf 0.38} & 169.02 
                          & 0.70 & 0.48 & 0.35 & {\bf 221.25} \\ 
                          
            \bf Communication & (0.19) & (0.23) & (0.17) & (1.59) 
                             & (0.14) & (0.11) & (0.13) & (105.65) 
                             & (0.11) & (0.13) & (0.13) & (51.63)\\           
            \hline
            \bf Emergent & 0.77 & {\bf 0.43} & {\bf 0.27} & {\bf 11.41} 
                         & {\bf 0.81} & 0.48 & 0.36 & {\bf 186.50} 
                         & {\bf 0.74} & 0.45 & 0.34 & 197.75 \\ 
            \bf Global-Centric & (0.16) & (0.14) & (0.08) & (2.05) 
                               & (0.11) & (0.13) & (0.12) & (101.67) 
                               & (0.10) & (0.17) & (0.14) & (67.33)\\          
            \hline
            \bf Emergent & 0.64 & 0.33 & 0.15 & 4.81 
                         & 0.74 & {\bf 0.52} & 0.33 & 131.68 
                         & 0.67 & 0.49 & {\bf 0.38} & 140.15 \\ 
            \bf Self-Centric & (0.21) & (0.17) & (0.14) & (0.87) 
                             & (0.14) & (0.13) & (0.12) & (94.76) 
                             & (0.11) & (0.14) & (0.15) & (40.42)\\      
            \hline 
        \end{tabular} 
        }
    \end{center}
    \label{table:c_levels}
\end{table*}

%% file: content/results-graphs4.tex
\begin{figure}[!htbp]
    \centering
    \begin{minipage}[b]{0.49\textwidth}
        \includegraphics[width=\textwidth,height=3.75cm]{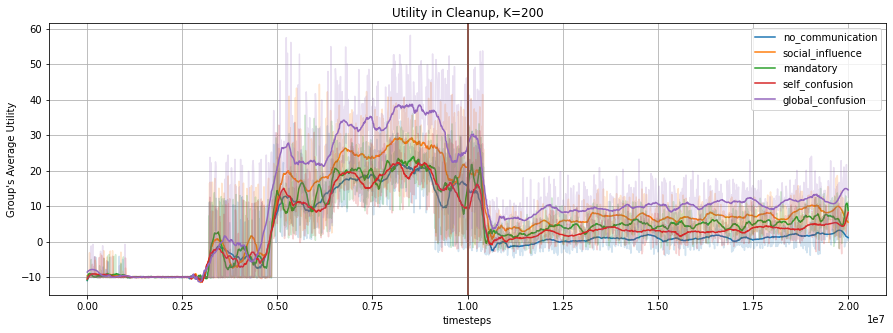}
        \caption{Average utility for \(K = 200\) in the Cleanup environment domain before and after a perturbation occurs.}
        \label{fig:cleanup-training}
    \end{minipage}\hfill
    \begin{minipage}[b]{0.49\textwidth}
        \includegraphics[width=\textwidth,height=3.75cm]{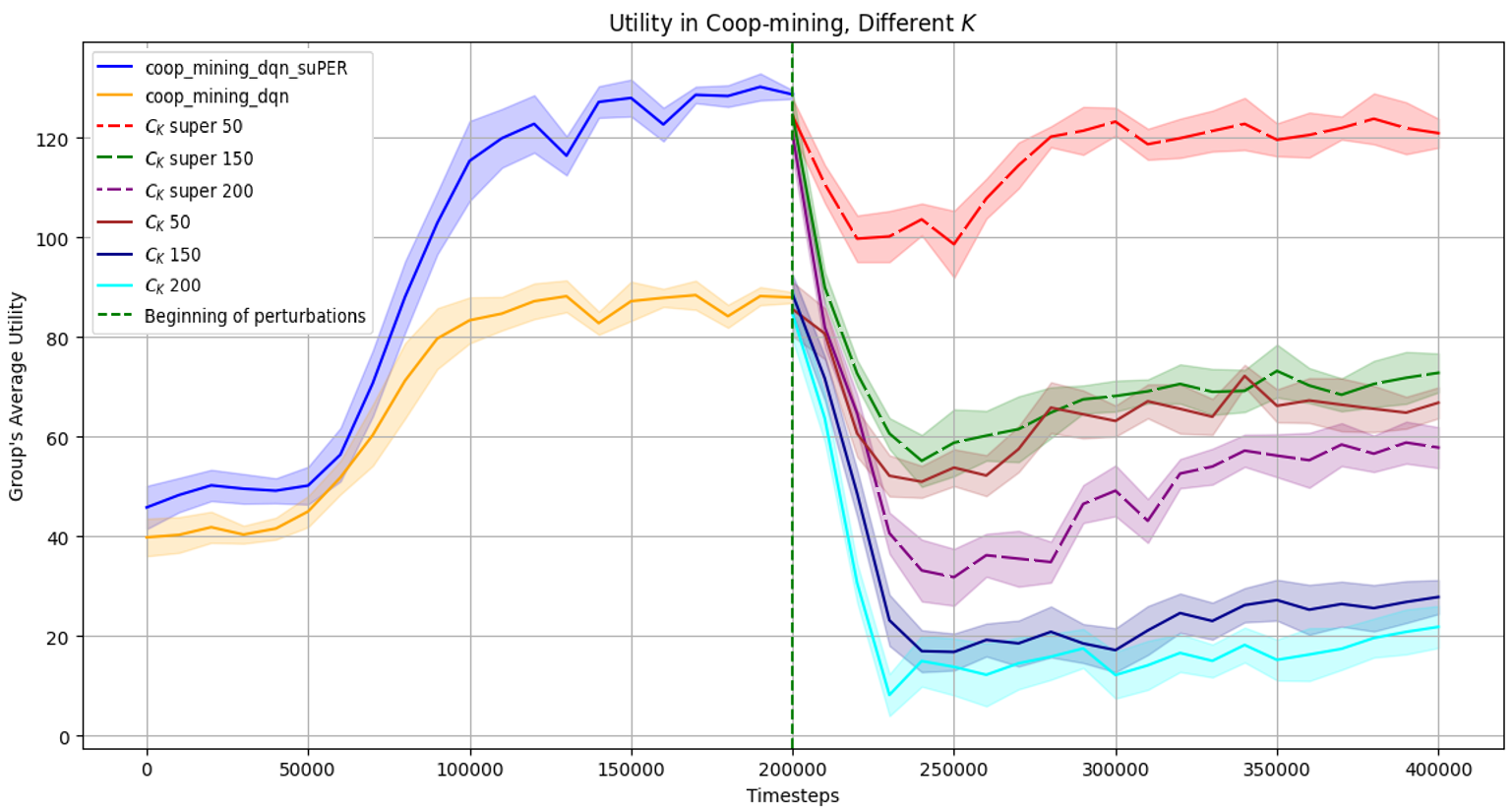}
        \caption{Average utility for different perturbation magnitudes in the Coop-mining environment domain before and after a perturbation occurs.}
        \label{fig:coop-training}
    \end{minipage}

    \vspace{0.15cm}

    \begin{minipage}[b]{0.49\textwidth}
        \includegraphics[width=\textwidth,height=4.8cm]{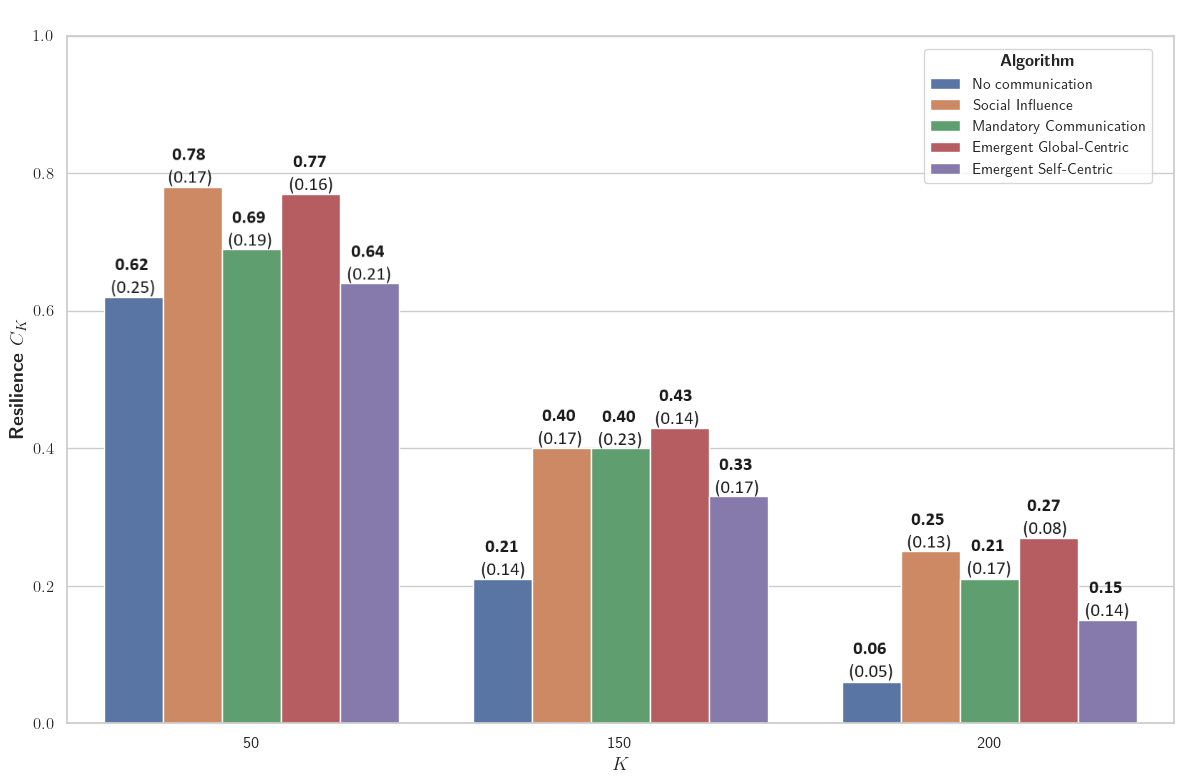}
        \caption{Cleanup environment resilience.}
        \label{fig:cleanup-compare}
    \end{minipage}\hfill
    \begin{minipage}[b]{0.49\textwidth}
        \includegraphics[width=\textwidth,height=4.8cm]{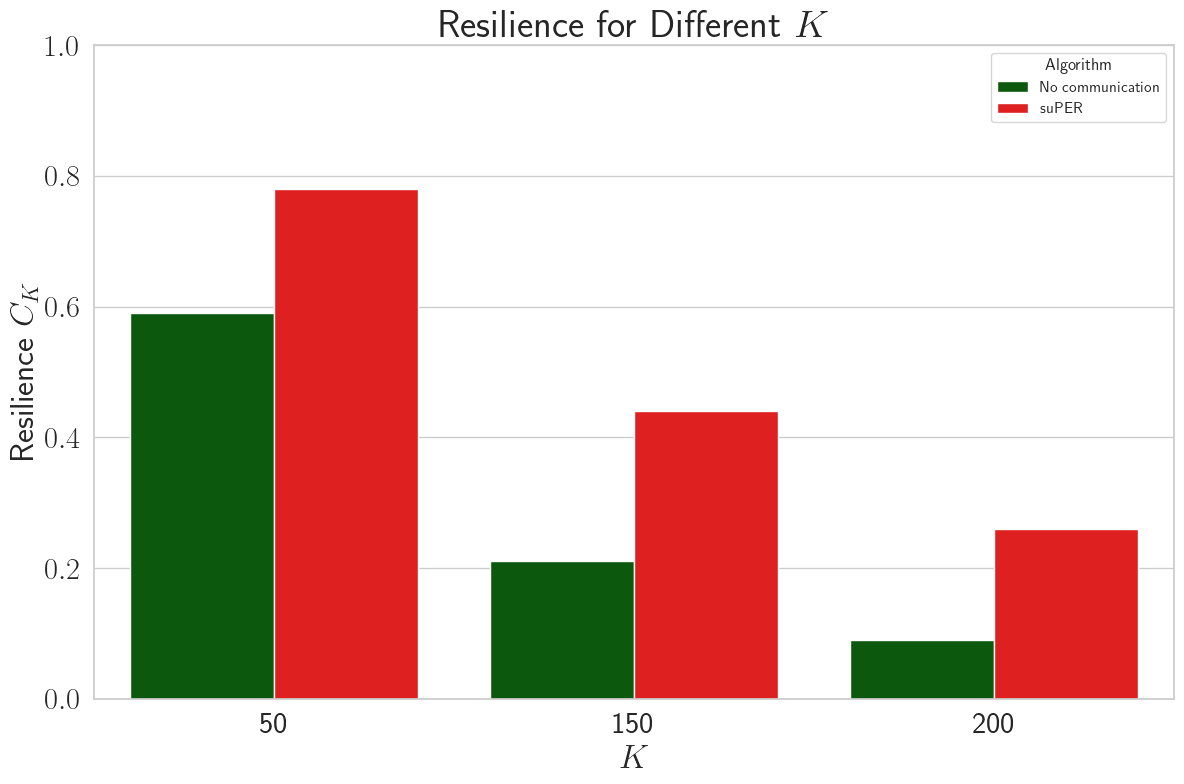}
        \caption{Coop-mining environment resilience.}
        \label{fig:coop-compare}
    \end{minipage}
\end{figure}

%% file: main.bbl
\begin{thebibliography}{34}
\providecommand{\natexlab}[1]{#1}
\providecommand{\url}[1]{\texttt{#1}}
\expandafter\ifx\csname urlstyle\endcsname\relax
  \providecommand{\doi}[1]{DOI: #1}\else
  \providecommand{\doi}{DOI: \begingroup \urlstyle{rm}\Url}\fi

\bibitem[Azran et~al.(2024)Azran, Danesh, Albrecht, and Keren]{azranContextualPreplanningReward2024}
Guy Azran, Mohamad~H. Danesh, Stefano~V. Albrecht, and Sarah Keren.
\newblock Contextual {{Pre-planning}} on {{Reward Machine Abstractions}} for {{Enhanced Transfer}} in {{Deep Reinforcement Learning}}.
\newblock \emph{Proceedings of the AAAI Conference on Artificial Intelligence}, 38\penalty0 (10):\penalty0 10953--10961, March 2024.
\newblock ISSN 2374-3468.
\newblock \doi{10.1609/aaai.v38i10.28970}.

\bibitem[Bellman(1957)]{bellmanMarkovianDecisionProcess1957}
Richard Bellman.
\newblock A {{Markovian Decision Process}}.
\newblock \emph{Indiana University Mathematics Journal}, 6\penalty0 (4):\penalty0 679--684, 1957.
\newblock ISSN 0022-2518.
\newblock \doi{10.1512/iumj.1957.6.56038}.

\bibitem[Bou~Ammar et~al.(2014)Bou~Ammar, Eaton, Taylor, Mocanu, Driessens, Weiss, and Tuyls]{ammarAutomatedMeasureMDPSimilarity2014}
Haitham Bou~Ammar, Eric Eaton, Matthew~E. Taylor, Decebal~C. Mocanu, Kurt Driessens, Gerhard Weiss, and Karl Tuyls.
\newblock An automated measure of {{MDP}} similarity for transfer in reinforcement learning.
\newblock In \emph{Machine Learning for Interactive Systems Workshop (WS‑14‑07), 28th AAAI Conference on Artificial Intelligence}, pp.\  31--37, Québec City, QC, Canada, 2014. AAAI Press.

\bibitem[Christianos et~al.(2020)Christianos, Sch{\"a}fer, and Albrecht]{christianosSharedExperienceActorCritic2020}
Filippos Christianos, Lukas Sch{\"a}fer, and Stefano Albrecht.
\newblock Shared {{Experience Actor-Critic}} for {{Multi-Agent Reinforcement Learning}}.
\newblock In \emph{Advances in {{Neural Information Processing Systems}}}, volume~33, pp.\  10707--10717. Curran Associates, Inc., 2020.

\bibitem[Dobrushin(1970)]{dobrushinPrescribingSystemRandom1970}
R.~L. Dobrushin.
\newblock Prescribing a {{System}} of {{Random Variables}} by {{Conditional Distributions}}.
\newblock \emph{Theory of Probability \& Its Applications}, 15\penalty0 (3):\penalty0 458--486, January 1970.
\newblock ISSN 0040-585X, 1095-7219.
\newblock \doi{10.1137/1115049}.

\bibitem[Ferns et~al.(2004)Ferns, Panangaden, and Precup]{fernsMetricsFiniteMarkovDecision2004}
Norman Ferns, Prakash Panangaden, and Doina Precup.
\newblock Metrics for finite markov decision processes.
\newblock In \emph{Proceedings of the 20th Conference on Uncertainty in Artificial Intelligence (UAI)}, pp.\  162--169, Banff, AB, Canada, 2004. AUAI Press.

\bibitem[Foerster et~al.(2016{\natexlab{a}})Foerster, Assael, de~Freitas, and Whiteson]{Foerster2016LearningTC}
Jakob Foerster, Yannis~M. Assael, Nando de~Freitas, and Shimon Whiteson.
\newblock Learning to communicate with deep multi-agent reinforcement learning.
\newblock In \emph{Advances in Neural Information Processing Systems (NeurIPS)}, pp.\  2137--2145, 2016{\natexlab{a}}.

\bibitem[Foerster et~al.(2016{\natexlab{b}})Foerster, Assael, de~Freitas, and Whiteson]{foerster_learning_2016}
Jakob~N. Foerster, Yannis~M. Assael, Nando de~Freitas, and Shimon Whiteson.
\newblock Learning to {Communicate} with {Deep} {Multi}-{Agent} {Reinforcement} {Learning}, May 2016{\natexlab{b}}.
\newblock URL \url{http://arxiv.org/abs/1605.06676}.
\newblock arXiv:1605.06676 [cs].

\bibitem[Gerstgrasser et~al.(2023)Gerstgrasser, Danino, and Keren]{gerstgrasserSelectivelySharingExperiences2023}
Matthias Gerstgrasser, Tom Danino, and Sarah Keren.
\newblock Selectively {{Sharing Experiences Improves Multi-Agent Reinforcement Learning}}.
\newblock \emph{Advances in Neural Information Processing Systems}, 36:\penalty0 59543--59565, December 2023.

\bibitem[Honhaga \& Szabo(2024)Honhaga and Szabo]{honhaga2024simulation}
Ishan Honhaga and Claudia Szabo.
\newblock A simulation and experimentation architecture for resilient cooperative multiagent reinforcement learning models operating in contested and dynamic environments.
\newblock \emph{SIMULATION}, pp.\  00375497241232432, 2024.

\bibitem[Jaques et~al.(2019{\natexlab{a}})Jaques, Lazaridou, Hughes, Gulcehre, Ortega, Strouse, Leibo, and de~Freitas]{Jaques2019SocialIA}
Natasha Jaques, Angeliki Lazaridou, Edward Hughes, Caglar Gulcehre, Pedro Ortega, DJ~Strouse, Joel~Z. Leibo, and Nando de~Freitas.
\newblock Social influence as intrinsic motivation for multi-agent deep reinforcement learning.
\newblock In \emph{Proceedings of the 36th International Conference on Machine Learning (ICML)}, volume~97, pp.\  3040--3049. PMLR, 2019{\natexlab{a}}.

\bibitem[Jaques et~al.(2019{\natexlab{b}})Jaques, Lazaridou, Hughes, Gulcehre, Ortega, Strouse, Leibo, and de~Freitas]{jaques_social_2019}
Natasha Jaques, Angeliki Lazaridou, Edward Hughes, Caglar Gulcehre, Pedro~A. Ortega, D.~J. Strouse, Joel~Z. Leibo, and Nando de~Freitas.
\newblock Social {Influence} as {Intrinsic} {Motivation} for {Multi}-{Agent} {Deep} {Reinforcement} {Learning}, June 2019{\natexlab{b}}.
\newblock URL \url{http://arxiv.org/abs/1810.08647}.
\newblock arXiv:1810.08647 [cs, stat].

\bibitem[Jordan \& Trott(2020)Jordan and Trott]{jordanScoreNormalizationMultiEnv2020}
Spencer Jordan and Alex Trott.
\newblock Normalization of returns across multiple environments.
\newblock In \emph{Proceedings of Machine Learning Research}, pp.\  1--12, 2020.
\newblock To appear; fill in exact volume \& pages if available.

\bibitem[Leibo et~al.(2021)Leibo, nez Guzm\'an, Vezhnevets, Agapiou, Sunehag, Koster, Matyas, Beattie, Mordatch, and Graepel]{leibo2021meltingpot}
Joel~Z. Leibo, Edgar~Du\'e\ nez Guzm\'an, Alexander~Sasha Vezhnevets, John~P. Agapiou, Peter Sunehag, Raphael Koster, Jayd Matyas, Charles Beattie, Igor Mordatch, and Thore Graepel.
\newblock Scalable evaluation of multi-agent reinforcement learning with melting pot.
\newblock In \emph{International conference on machine learning}. PMLR, 2021.
\newblock \doi{10.48550/arXiv.2107.06857}.
\newblock URL \url{https://doi.org/10.48550/arXiv.2107.06857}.

\bibitem[Liang \& Li(2020)Liang and Li]{liang_parallel_2020}
Yongyuan Liang and Bangwei Li.
\newblock Parallel {Knowledge} {Transfer} in {Multi}-{Agent} {Reinforcement} {Learning}, March 2020.
\newblock URL \url{http://arxiv.org/abs/2003.13085}.
\newblock arXiv:2003.13085 [cs].

\bibitem[Littman(1994)]{littmanMarkovGamesFramework1994a}
Michael~L. Littman.
\newblock Markov games as a framework for multi-agent reinforcement learning.
\newblock In William~W. Cohen and Haym Hirsh (eds.), \emph{Machine {{Learning Proceedings}} 1994}, pp.\  157--163. Morgan Kaufmann, San Francisco (CA), January 1994.
\newblock ISBN 978-1-55860-335-6.
\newblock \doi{10.1016/B978-1-55860-335-6.50027-1}.

\bibitem[Lowe et~al.(2020)Lowe, Wu, Tamar, Harb, Abbeel, and Mordatch]{lowe_multi-agent_2020}
Ryan Lowe, Yi~Wu, Aviv Tamar, Jean Harb, Pieter Abbeel, and Igor Mordatch.
\newblock Multi-{Agent} {Actor}-{Critic} for {Mixed} {Cooperative}-{Competitive} {Environments}, March 2020.
\newblock URL \url{http://arxiv.org/abs/1706.02275}.
\newblock arXiv:1706.02275 [cs].

\bibitem[Mahajan et~al.(2019)Mahajan, Rashid, Samvelyan, and Whiteson]{mahajanMAVENMultiAgentVariational2019}
Anuj Mahajan, Tabish Rashid, Mikayel Samvelyan, and Shimon Whiteson.
\newblock {{MAVEN}}: {{Multi-Agent Variational Exploration}}.
\newblock In \emph{Advances in {{Neural Information Processing Systems}}}, volume~32. Curran Associates, Inc., 2019.

\bibitem[Mnih et~al.(2015)Mnih, Kavukcuoglu, Silver, Rusu, Veness, Bellemare, Graves, Riedmiller, Fidjeland, Ostrovski, Petersen, Beattie, Sadik, Antonoglou, King, Kumaran, Wierstra, Legg, and Hassabis]{mnihHumanlevelControlDeep2015}
Volodymyr Mnih, Koray Kavukcuoglu, David Silver, Andrei~A. Rusu, Joel Veness, Marc~G. Bellemare, Alex Graves, Martin Riedmiller, Andreas~K. Fidjeland, Georg Ostrovski, Stig Petersen, Charles Beattie, Amir Sadik, Ioannis Antonoglou, Helen King, Dharshan Kumaran, Daan Wierstra, Shane Legg, and Demis Hassabis.
\newblock Human-level control through deep reinforcement learning.
\newblock \emph{Nature}, 518\penalty0 (7540):\penalty0 529--533, February 2015.
\newblock ISSN 1476-4687.
\newblock \doi{10.1038/nature14236}.

\bibitem[Mnih et~al.(2016)Mnih, Badia, Mirza, Graves, Lillicrap, Harley, Silver, and Kavukcuoglu]{mnih_asynchronous_2016}
Volodymyr Mnih, Adrià~Puigdomènech Badia, Mehdi Mirza, Alex Graves, Timothy~P. Lillicrap, Tim Harley, David Silver, and Koray Kavukcuoglu.
\newblock Asynchronous {Methods} for {Deep} {Reinforcement} {Learning}, June 2016.
\newblock URL \url{http://arxiv.org/abs/1602.01783}.
\newblock arXiv:1602.01783 [cs].

\bibitem[Padakandla(2022)]{padakandla_survey_2022}
Sindhu Padakandla.
\newblock A {Survey} of {Reinforcement} {Learning} {Algorithms} for {Dynamically} {Varying} {Environments}.
\newblock \emph{ACM Computing Surveys}, 54\penalty0 (6):\penalty0 1--25, July 2022.
\newblock ISSN 0360-0300, 1557-7341.
\newblock \doi{10.1145/3459991}.
\newblock URL \url{http://arxiv.org/abs/2005.10619}.
\newblock arXiv:2005.10619 [cs, stat].

\bibitem[Pattanaik et~al.(2017)Pattanaik, Tang, Liu, Bommannan, and Chowdhary]{pattanaik_robust_2017}
Anay Pattanaik, Zhenyi Tang, Shuijing Liu, Gautham Bommannan, and Girish Chowdhary.
\newblock Robust {Deep} {Reinforcement} {Learning} with {Adversarial} {Attacks}, December 2017.
\newblock URL \url{http://arxiv.org/abs/1712.03632}.
\newblock arXiv:1712.03632 [cs].

\bibitem[Phan et~al.(2020)Phan, Gabor, Sedlmeier, Ritz, Kempter, Klein, Sauer, Schmid, Wieghardt, Zeller, and Linnhoff-Popien]{phan_learning_2020}
Thomy Phan, Thomas Gabor, Andreas Sedlmeier, Fabian Ritz, Bernhard Kempter, Cornel Klein, Horst Sauer, Reiner Schmid, Jan Wieghardt, Marc Zeller, and Claudia Linnhoff-Popien.
\newblock Learning and {Testing} {Resilience} in {Cooperative} {Multi}-{Agent} {Systems}.
\newblock \emph{New Zealand}, 2020.

\bibitem[Portelas et~al.(2020)Portelas, Colas, Weng, Hofmann, and Oudeyer]{portelas_automatic_2020}
Rémy Portelas, Cédric Colas, Lilian Weng, Katja Hofmann, and Pierre-Yves Oudeyer.
\newblock Automatic {Curriculum} {Learning} {For} {Deep} {RL}: {A} {Short} {Survey}, May 2020.
\newblock URL \url{http://arxiv.org/abs/2003.04664}.
\newblock arXiv:2003.04664 [cs, stat].

\bibitem[Qian et~al.(2019)Qian, Wu, Wang, Zhu, and Zhang]{qianSurveyReinforcementLearning2019}
Yichen Qian, Jun Wu, Rui Wang, Fusheng Zhu, and Wei Zhang.
\newblock Survey on {{Reinforcement Learning Applications}} in {{Communication Networks}}.
\newblock \emph{Journal of Communications and Information Networks}, 4\penalty0 (2):\penalty0 30--39, June 2019.
\newblock ISSN 2509-3312.
\newblock \doi{10.23919/JCIN.2019.8917870}.

\bibitem[Rashid et~al.(2018)Rashid, Samvelyan, de~Witt, Farquhar, Foerster, and Whiteson]{rashid_qmix_2018}
Tabish Rashid, Mikayel Samvelyan, Christian~Schroeder de~Witt, Gregory Farquhar, Jakob Foerster, and Shimon Whiteson.
\newblock {QMIX}: {Monotonic} {Value} {Function} {Factorisation} for {Deep} {Multi}-{Agent} {Reinforcement} {Learning}, June 2018.
\newblock URL \url{http://arxiv.org/abs/1803.11485}.
\newblock arXiv:1803.11485 [cs, stat].

\bibitem[Saulnier et~al.(2017)Saulnier, Saldana, Prorok, Pappas, and Kumar]{saulnier_resilient_2017}
Kelsey Saulnier, David Saldana, Amanda Prorok, George~J. Pappas, and Vijay Kumar.
\newblock Resilient {Flocking} for {Mobile} {Robot} {Teams}.
\newblock \emph{IEEE Robotics and Automation Letters}, 2\penalty0 (2):\penalty0 1039--1046, April 2017.
\newblock ISSN 2377-3766, 2377-3774.
\newblock \doi{10.1109/LRA.2017.2655142}.
\newblock URL \url{http://ieeexplore.ieee.org/document/7822915/}.

\bibitem[Schaul et~al.(2016)Schaul, Quan, Antonoglou, and Silver]{schaul_prioritized_2016}
Tom Schaul, John Quan, Ioannis Antonoglou, and David Silver.
\newblock Prioritized {Experience} {Replay}, February 2016.
\newblock URL \url{http://arxiv.org/abs/1511.05952}.
\newblock arXiv:1511.05952 [cs].

\bibitem[Song et~al.(2016)Song, Gao, Wang, and An]{songMeasuringDistanceFinite2016a}
Jinhua Song, Yang Gao, Hao Wang, and Bo~An.
\newblock Measuring the {{Distance Between Finite Markov Decision Processes}}.
\newblock In \emph{Proceedings of the 2016 {{International Conference}} on {{Autonomous Agents}} \& {{Multiagent Systems}}}, {{AAMAS}} '16, pp.\  468--476, Richland, SC, May 2016. {International Foundation for Autonomous Agents and Multiagent Systems}.
\newblock ISBN 978-1-4503-4239-1.

\bibitem[Vinitsky et~al.(2019)Vinitsky, Jaques, Leibo, Castenada, and Hughes]{SSDOpenSource}
Eugene Vinitsky, Natasha Jaques, Joel Leibo, Antonio Castenada, and Edward Hughes.
\newblock An open source implementation of sequential social dilemma games.
\newblock \url{https://github.com/eugenevinitsky/sequential_social_dilemma_games/issues/182}, 2019.
\newblock GitHub repository.

\bibitem[Vinitsky et~al.(2020)Vinitsky, Du, Parvate, Jang, Abbeel, and Bayen]{vinitsky_robust_2020}
Eugene Vinitsky, Yuqing Du, Kanaad Parvate, Kathy Jang, Pieter Abbeel, and Alexandre Bayen.
\newblock Robust {Reinforcement} {Learning} using {Adversarial} {Populations}, September 2020.
\newblock URL \url{http://arxiv.org/abs/2008.01825}.
\newblock arXiv:2008.01825 [cs, stat].

\bibitem[Xu et~al.(2012)Xu, Rao, and Bu]{xu_url_2012}
Cheng-Zhong Xu, Jia Rao, and Xiangping Bu.
\newblock {URL}: {A} unified reinforcement learning approach for autonomic cloud management.
\newblock \emph{Journal of Parallel and Distributed Computing}, 72\penalty0 (2):\penalty0 95--105, February 2012.
\newblock ISSN 07437315.
\newblock \doi{10.1016/j.jpdc.2011.10.003}.
\newblock URL \url{https://linkinghub.elsevier.com/retrieve/pii/S0743731511001924}.

\bibitem[Zhang et~al.(2017)Zhang, Zhang, and Gupta]{zhangResilientRobotsConcept2017}
Tan Zhang, Wenjun Zhang, and Madan Gupta.
\newblock Resilient {{Robots}}: {{Concept}}, {{Review}}, and {{Future Directions}}.
\newblock \emph{Robotics}, 6\penalty0 (4):\penalty0 22, September 2017.
\newblock ISSN 2218-6581.
\newblock \doi{10.3390/robotics6040022}.

\bibitem[Zhu et~al.(2023)Zhu, Lin, Jain, and Zhou]{zhu_transfer_2023}
Zhuangdi Zhu, Kaixiang Lin, Anil~K. Jain, and Jiayu Zhou.
\newblock Transfer {Learning} in {Deep} {Reinforcement} {Learning}: {A} {Survey}, July 2023.
\newblock URL \url{http://arxiv.org/abs/2009.07888}.
\newblock arXiv:2009.07888 [cs, stat].

\end{thebibliography}
